\documentclass[conference,10pt]{IEEEtran}
\usepackage{amsmath,amssymb,amsfonts}
\usepackage{graphicx}
\usepackage{subfig}
\usepackage[hyphens]{url}
\usepackage{multirow}
\usepackage{booktabs}
\def\eg{\textit{e.g}. } 
\def\ie{\textit{i.e}. }

\def\wrt{w.r.t. } 
\def\paperpage{\url{http://wavelab.at/sources/Drozdowski20a/}}
\newcommand{\NA}{---}

\makeatletter
\newcommand*\titleheader[1]{\gdef\@titleheader{#1}}
\AtBeginDocument{%
  \let\st@red@title\@title
  \def\@title{%
    \bgroup\normalfont\footnotesize\centering\@titleheader\par\egroup
    \vskip0.25em\st@red@title}
}
\makeatother

\titleheader{Submitted to European Signal Processing Conference (EUSIPCO) -- special session on bias in biometrics}

\usepackage{hyperref}
\hypersetup{
pdfauthor={P. Drozdowski, B. Prommegger, G. Wimmer, R. Schraml, C. Rathgeb, A. Uhl, C. Busch},
pdfsubject={Biometrics},
pdftitle={Demographic Bias: A Challenge for Fingervein Recognition Systems?},
pdfkeywords={Biometrics, Fingervein Recognition, Bias},
bookmarks=true,
bookmarksopen=true,
}

\title{Demographic Bias: \\ A Challenge for Fingervein Recognition Systems?}

\author{P. Drozdowski\textsuperscript{$*$}, B. Prommegger\textsuperscript{$\dagger$}, G. Wimmer\textsuperscript{$\dagger$}, R. Schraml\textsuperscript{$\dagger$}, C. Rathgeb\textsuperscript{$*$}, A. Uhl\textsuperscript{$\dagger$}, and C. Busch\textsuperscript{$*$} \vspace{0.25cm} \\  
\textsuperscript{$*$} da/sec - Biometrics and Internet Security Research Group, Hochschule Darmstadt, Germany \\ 
\textsuperscript{$\dagger$} Multimedia Signal Processing and Security Lab, University of Salzburg, Austria \vspace{0.15cm} \\
{\tt\small \{pawel.drozdowski,christian.rathgeb,christoph.busch\}@h-da.de} \\
{\tt\small \{bprommeg,gwimmer,rschraml,uhl\}@cosy.sbg.ac.at}
}

\begin{document}

\maketitle

\begin{abstract}
Recently, concerns regarding potential biases in the underlying algorithms of many automated systems (including biometrics) have been raised. In this context, a biased algorithm produces statistically different outcomes for different groups of individuals based on certain (often protected by anti-discrimination legislation) attributes such as sex and age. While several preliminary studies investigating this matter for facial recognition algorithms do exist, said topic has not yet been addressed for vascular biometric characteristics. Accordingly, in this paper, several popular types of recognition algorithms are benchmarked to ascertain the matter for fingervein recognition. The experimental evaluation suggests lack of bias for the tested algorithms, although future works with larger datasets are needed to validate and confirm those preliminary results.
\end{abstract}


\section{Introduction}
\label{sec:introduction}
Automated systems (including biometrics) are increasingly used in decision making processes within various domains, some of which have traditionally enjoyed strong anti-discrimination legislation protection (see \eg \cite{EU-AntiDiscrimination-2018}). 

\begin{figure}[!ht]
\centering
\includegraphics[width=\linewidth]{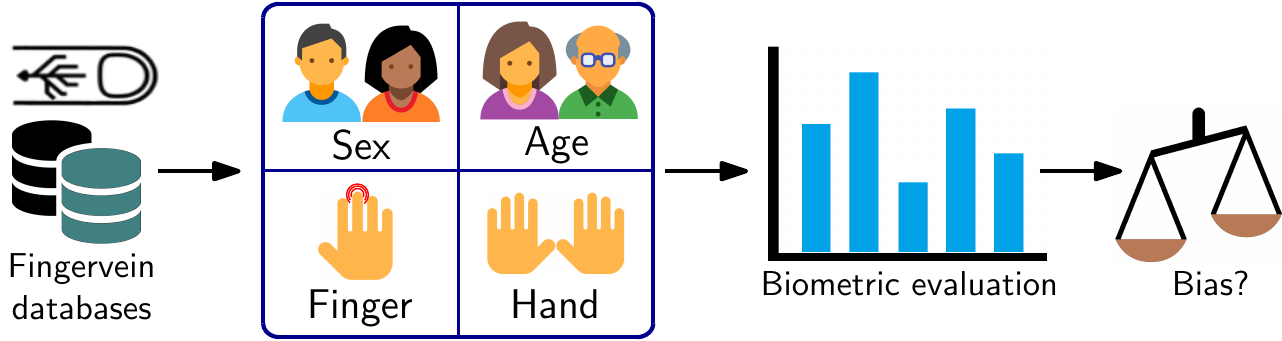}
\caption{Overview of the conducted experiments}
\label{fig:overview}
\end{figure}

In recent years, substantial media coverage of systemic biases inherent to several such systems have been reported and hotly debated. In this context, a biased algorithm produces statistically different outcomes (decisions) for different groups of individuals, \eg based on sex, age, and ethnicity \cite{nature2016editorial}. For biometric recognition specifically, it means that the score distributions and therefore the chances of false positives and/or false negatives may vary across the groups. This, in consequence, may impact the outcomes on the system/application level -- for example, one recent study claimed disproportionally high arrest and search rates for certain groups based on decisions made by automatic facial recognition software \cite{garvie2016perpetual}. 

Although some studies have approached bias measurements and ensuring fairness in various machine learning contexts (see \eg \cite{kilbertus2017avoiding} and \cite{hardt2016equality}), for computer vision and biometrics in particular, this remains a nascent field of research. In \cite{phillips2011other}, it was reported that facial recognition algorithms tend to exhibit higher biometric performance with individuals from ethnic groups corresponding to the area of development of the algorithm; presumably due to training data availability. In \cite{klare2012demographics} and \cite{buolamwini2018gender}, some facial biometrics algorithms were shown to exhibit lower recognition and classification performances for certain groups of individuals (in particular women and non-white people), whereas \cite{Grother-NIST-FRVTBias-2019} conducted a large-scale benchmark of commercial and academic algorithms controlling for various demographic (and other) attributes. Proof-of-concept studies into bias mitigation for facial soft biometric classification using neural networks were presented in \eg \cite{das2018mitigating} and \cite{ryu2017inclusivefacenet}. 

While facial recognition is certainly the most widely covered and discussed biometric characteristic recently, also with some existing preliminary studies in the context of bias and fairness, this topic remains even less explored for other biometric characteristics. In general, algorithmic bias is considered (by some influential researchers) to be one of the, as of yet unresolved, challenges in biometric systems \cite{Ross-OpenProblems-2019}. For fingervein specifically, a small study evaluating the biometric performance \wrt sex and age of the subjects (as a part of a paper presenting a new dataset) has been conducted in \cite{Kauba-PLUS-2018}. Intuitively, the demographic covariates could be proxies for certain anatomical features which might influence the performance of fingervein recognition systems (\eg the thickness of the finger). Otherwise, currently no studies benchmarking the potential biases in fingervein algorithms have been reported in the scientific literature. In this paper, a benchmark is conducted to address the following two questions, see figure \ref{fig:overview}:

\begin{itemize}
\item Do score distributions computed by fingervein recognition algorithms on disjoint groups of data instances (\ie based on metadata attributes, such as subject sex, age, and others) exhibit statistically significant differences?
\item Do these results persist across fundamentally different types of fingervein recognition algorithms?
\end{itemize}

The remainder of this paper is organised as follows: the experimental setup is described in section \ref{sec:experimentalsetup}. The results of the evaluation are presented in section \ref{sec:evaluation}, while concluding remarks and a summary are given in section \ref{sec:conclusion}.

\section{Experimental Setup}
\label{sec:experimentalsetup}
In this section, the used datasets (subsection \ref{subsec:datasets}) and data processing pipelines (subsection \ref{subsec:algorithms}) are presented.

\subsection{Datasets}
\label{subsec:datasets}
\begin{figure}[!ht]
\vspace{-0.75cm}
\centering
\subfloat[MMCBNU]{\includegraphics[height=3.1cm]{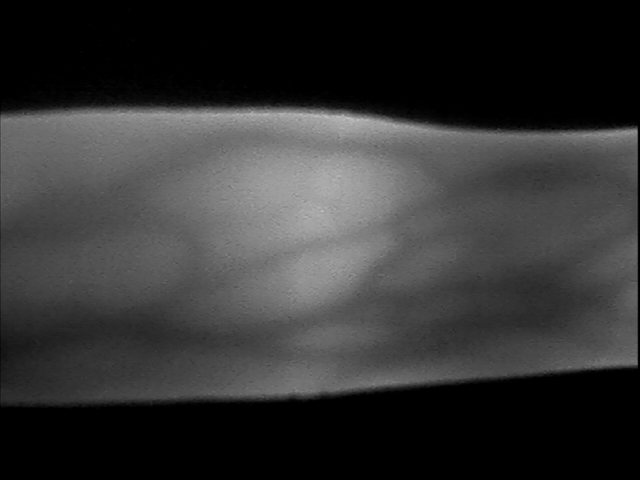}} \hfill
\subfloat[PLUS]{\includegraphics[height=3.1cm]{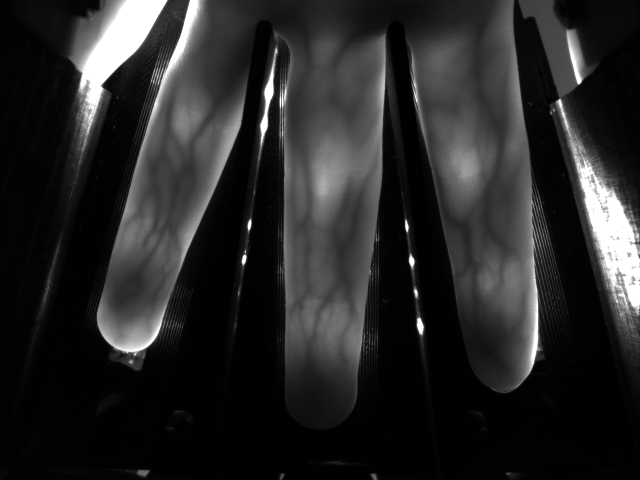}} \hfill
\subfloat[UTFVP]{\includegraphics[height=1.55cm]{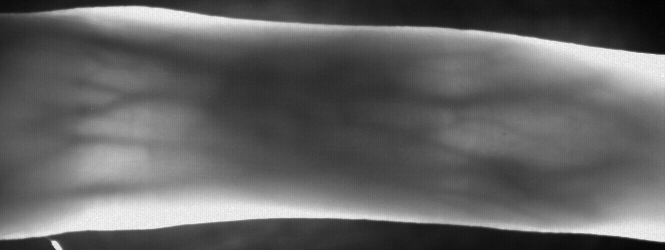}} \hfill
\subfloat[VERA]{\includegraphics[height=1.55cm]{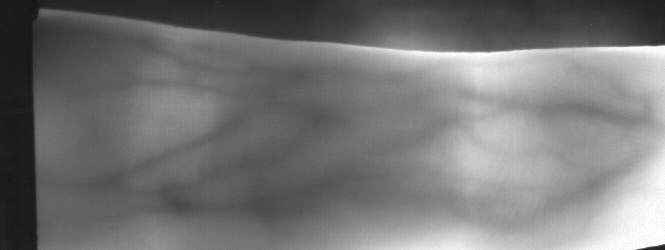}} \hfill
\caption{Example images from the chosen datasets}
\label{fig:exampleimages}
\vspace{-0.3cm}
\end{figure}

Four publicly available fingervein databases with metadata labels were used. They are listed in table \ref{table:datasets}, while example images are shown in figure \ref{fig:exampleimages}. The datasets were chosen based on the presence of the metadata information (sex and age), as well as the presence of samples from both hands and different fingers.

\begin{table}[!ht]
\centering
\caption{Summary of the chosen datasets}
\label{table:datasets}
\resizebox{0.7\linewidth}{!}{
\begin{tabular}{lllllll}
\toprule
Database & Subjects & Instances & Samples \\ 
\midrule
MMCBNU \cite{Lu-MMCBNU-2013} & 100 & 600 & 6000 \\
PLUS \cite{Kauba-PLUS-2018,Sequeira18a} & 78 & 468 & 2340 \\
UTFVP \cite{Ton-UTFVP-2013} & 60 & 360 & 1440 \\
VERA \cite{Vanoni-VERA-2014} & 110 & 220 & 440 \\
\bottomrule 
\end{tabular}
}
\end{table}

\subsection{Feature Types}
\label{subsec:featureTypes}
All experiments are executed using four fundamentally different types of vein recognition schemes:
\begin{enumerate}
	\item \textbf{Vein pattern}: The first two of the used techniques, Maximum Curvature (MC \cite{BMiura07a}) and Principal Curvature (PC \cite{BChoi09b}), aim to separate the vein pattern from the background resulting in a binary image.
	\item \textbf{Keypoints}: In contrast to the vein pattern techniques, key-point based techniques try to use information from the most discriminative points as well as considering the neighborhood and context information of these points by extracting key-points and assigning a descriptor to each key-point. We used a Scale-Invariant Feature Transform (SIFT \cite{Lowe99a}) based technique with additional key-point filtering. All details of this method are described in \cite{Kauba14a}.
	\item \textbf{Texture}: The used approach is an adapted version of \cite{BFeng2016a}. It combines Log Gabor convolution magnitude and local binary patterns (LBP \cite{OtherOjala96a}).
	\item \textbf{Deep learning}: LightCNN with triplet loss (LCNN \cite{Xie19}) is a small neural network which is trained using the triplet loss, a special loss function that enables the identification of subjects that were not included in the training set. By using a more advanced selection of the triplets (input images) for training (hard triplet online selection) and by omitting the supervised discrete hashing of the CNN outputs, the results could be improved \wrt \cite{Xie19}. It should be noted that the LCNN was trained according to its actual purpose, the recognition of individuals based on the captured vein images. If the net would be trained according to the chosen groups, such as sex or age, the results might look different.
\end{enumerate}

\begin{table}[!ht]
\centering
\caption{Biometric performance of the used methods}
\label{table:performance}
\resizebox{0.7\linewidth}{!}{
\begin{tabular}{llllll}
\toprule
\multirow{2}{*}{Dataset} & \multicolumn{5}{c}{EER (in \%)} \\
& LCNN & LBP & MC & PC & SIFT \\
\midrule
MMCBNU & 2.2 & 1.3 & 1.8 & 1.7 & 2.0 \\
PLUS & 4.5 & 3.6 & 0.5 & 0.2 & 0.8 \\
UTFVP & 7.0 & 1.5 & 0.2 & 0.4 & 1.5 \\
VERA & \NA & 3.2 & 1.8 & 2.3 & 2.6 \\
\bottomrule
\end{tabular}

}
\vspace{-0.3cm}
\end{table}

\subsection{Data Processing Pipelines}
\label{subsec:algorithms}
The finger vein recognition tool-chain consists of the following components:
(1) For \emph{finger region detection}, \emph{finger alignment} and \emph{ROI extraction} an implementation that is based on \cite{lu2013robust} is used.
(2) To improve the visibility of the vein pattern, \emph{High Frequency Emphasis Filtering} (HFE) \cite{BZhao09a}, \emph{Circular Gabor Filter} (CGF) \cite{BZhang09b}, and simple \emph{CLAHE} (local histogram equalisation) \cite{Zuiderveld94a} are used during \emph{pre-processing}.
(3a) For the simple vein pattern based feature methods, MC and PC, the binary feature images are compared using a correlation measure, calculated between the input images and in x- and y-direction shifted and rotated versions of the reference image as described in \cite{BMiura04a}.
(3b) The SIFT based method applies feature extraction and comparison as proposed in \cite{Kauba14a}, and
(3c) the LBP based approach as described in \cite{BFeng2016a} where the LDP features are replaced with LBP features from \cite{OtherOjala96a}, respectively. (3d) For LCNN, the ROIs extracted in (1) have been resized to the required input size of 256$\times$256. For separating training from input data, a 2-fold cross validation has been applied. Due to a limited number of samples, a training on the VERA data set was not possible. The biometric performance of the used methods is summarised in table \ref{table:performance}.\footnote{Supplementary files (\eg comparison scores) are available for download at \paperpage.} 

\section{Evaluation}
\label{sec:evaluation}
The conceptual overview of the conducted experiments is shown in figure \ref{fig:overview}. The descriptive statistics (mean -- $\mu$ and standard deviation -- $\sigma$) of the score distributions \wrt the metadata-based groupings are computed in subsections \ref{subsec:sex} to \ref{subsec:fingerhand}. Additionally, for comparison, table \ref{table:allstats} shows the same statistics for all template comparisons without any metadata-based grouping. The algorithms have different ranges of scores. In the context of the conducted bias benchmark, only intra-algorithm comparisons are meaningful -- therefore, score normalisation is not required. The results are only listed where available (\eg the VERA dataset only contains two samples per finger and therefore it is not possible to train a CNN using the triplet loss function).

\begin{table}[!ht]
\centering
\caption{Score distribution statistics without metadata-based grouping}
\label{table:allstats}
\resizebox{0.9\linewidth}{!}{
\begin{tabular}{llllll}
\toprule
\multirow{2}{*}{Dataset} & \multirow{2}{*}{Algorithm} & \multicolumn{2}{c}{Score $\mu$} & \multicolumn{2}{c}{Score $\sigma$} \\
& & Genuine & Impostor & Genuine & Impostor \\\midrule
MMCBNU & LCNN & 0.72477 & 0.26189 & 0.12129 & 0.07989 \\
& LBP & 0.84812 & 0.78781 & 0.02009 & 0.00805 \\
& MC & 0.27205 & 0.12055 & 0.04891 & 0.01730 \\
& PC & 0.41059 & 0.30588 & 0.02806 & 0.01654 \\
& SIFT & 0.39081 & 0.01510 & 0.16538 & 0.02434 \\
\midrule
PLUS & LCNN & 0.70367 & 0.25433 & 0.13798 & 0.08999 \\
& LBP & 0.80302 & 0.37204 & 0.16154 & 0.05899 \\
& MC & 0.25004 & 0.12464 & 0.03946 & 0.00874 \\
& PC & 0.41261 & 0.30209 & 0.03030 & 0.01361 \\
& SIFT & 0.35114 & 0.01119 & 0.14678 & 0.01196 \\
\midrule
UTFVP & LCNN & 0.69713 & 0.32198 & 0.11472 & 0.10833 \\
& LBP & 0.84664 & 0.81038 & 0.01344 & 0.00471 \\
& MC & 0.23834 & 0.11789 & 0.03854 & 0.00733 \\
& PC & 0.40156 & 0.28823 & 0.02816 & 0.01058 \\
& SIFT & 0.31269 & 0.00937 & 0.15406 & 0.01071 \\
\midrule
VERA & LBP & 0.81296 & 0.78861 & 0.01101 & 0.00417 \\
& MC & 0.24056 & 0.11351 & 0.04569 & 0.00947 \\
& PC & 0.38741 & 0.29499 & 0.03025 & 0.01156 \\
& SIFT & 0.23220 & 0.00562 & 0.14523 & 0.00822 \\
\bottomrule
\end{tabular}

}
\end{table}

\subsection{Sex}
\label{subsec:sex}
In this experiment, the data instances are grouped by the subject sex (male and female) with the genuine and impostor score distributions computed within the groups. Table \ref{table:sexstats} shows the descriptive statistics of those score distributions.

\begin{table}[!ht]
\centering
\caption{Score distribution statistics by subject sex}
\label{table:sexstats}
\resizebox{\linewidth}{!}{
\begin{tabular}{lllllll}
\toprule
\multirow{2}{*}{Dataset} & \multirow{2}{*}{Algorithm} & \multirow{2}{*}{Sex} & \multicolumn{2}{c}{Score $\mu$} & \multicolumn{2}{c}{Score $\sigma$} \\
& & & Genuine & Impostor & Genuine & Impostor \\\midrule
MMCBNU & LCNN & Male & 0.72368 & 0.25896 & 0.12250 & 0.07901 \\
& & Female & 0.73008 & 0.33211 & 0.11503 & 0.08697 \\
& LBP & Male & 0.84866 & 0.78734 & 0.02047 & 0.00826 \\
& & Female & 0.84547 & 0.79151 & 0.01791 & 0.00685 \\
& MC & Male & 0.27539 & 0.12060 & 0.04875 & 0.01733 \\
& & Female & 0.25575 & 0.12102 & 0.04635 & 0.01710 \\
& PC & Male & 0.41292 & 0.30600 & 0.02758 & 0.01652 \\
& & Female & 0.39922 & 0.30596 & 0.02764 & 0.01685 \\
& SIFT & Male & 0.39940 & 0.01577 & 0.16554 & 0.02541 \\
& & Female & 0.34890 & 0.01269 & 0.15805 & 0.01926 \\
\midrule
PLUS & LCNN & Male & 0.69420 & 0.25298 & 0.13924 & 0.08804 \\
& & Female & 0.71743 & 0.27548 & 0.13494 & 0.09270 \\
& LBP & Male & 0.79966 & 0.36896 & 0.16497 & 0.05888 \\
& & Female & 0.80798 & 0.37765 & 0.15622 & 0.05819 \\
& MC & Male & 0.24732 & 0.12233 & 0.04122 & 0.00846 \\
& & Female & 0.25398 & 0.12849 & 0.03640 & 0.00836 \\
& PC & Male & 0.41017 & 0.30100 & 0.03129 & 0.01399 \\
& & Female & 0.41617 & 0.30414 & 0.02844 & 0.01290 \\
& SIFT & Male & 0.35073 & 0.01192 & 0.15233 & 0.01227 \\
& & Female & 0.35173 & 0.01076 & 0.13833 & 0.01181 \\
\midrule
UTFVP & LCNN & Male & 0.71090 & 0.34391 & 0.11343 & 0.10297 \\
& & Female & 0.65934 & 0.34688 & 0.10964 & 0.11891 \\
& LBP & Male & 0.84929 & 0.81163 & 0.01259 & 0.00415 \\
& & Female & 0.83938 & 0.81127 & 0.01301 & 0.00477 \\
& MC & Male & 0.24182 & 0.11602 & 0.03774 & 0.00674 \\
& & Female & 0.22877 & 0.12381 & 0.03911 & 0.00769 \\
& PC & Male & 0.40795 & 0.28808 & 0.02460 & 0.01054 \\
& & Female & 0.38403 & 0.28944 & 0.02983 & 0.01064 \\
& SIFT & Male & 0.34616 & 0.00927 & 0.14795 & 0.01062 \\
& & Female & 0.22082 & 0.01078 & 0.13145 & 0.01148 \\
\midrule
VERA & LBP & Male & 0.81534 & 0.78890 & 0.01149 & 0.00416 \\
& & Female & 0.80881 & 0.78950 & 0.00869 & 0.00398 \\
& MC & Male & 0.24689 & 0.11319 & 0.04681 & 0.00946 \\
& & Female & 0.22948 & 0.11463 & 0.04140 & 0.00929 \\
& PC & Male & 0.39161 & 0.29269 & 0.02930 & 0.01118 \\
& & Female & 0.38007 & 0.30051 & 0.03050 & 0.01085 \\
& SIFT & Male & 0.26795 & 0.00549 & 0.15266 & 0.00823 \\
& & Female & 0.16964 & 0.00621 & 0.10518 & 0.00855 \\
\bottomrule
\end{tabular}

}
\end{table}

\begin{table}[!ht]
\centering
\caption{$Z$-scores summary for all the experiments}
\label{table:zstats}
\resizebox{0.75\linewidth}{!}{
\begin{tabular}{lllll}
\toprule
\multirow{2}{*}{Attribute} & \multicolumn{2}{c}{Z-score median} & \multicolumn{2}{c}{Z-score maximum} \\
& Genuine & Impostor & Genuine & Impostor \\\midrule
Sex & 0.23043 & 0.10282 & 0.63334 & 0.76144 \\
Age & 0.09125 & 0.08500 & 0.29284 & 0.31685 \\
Finger & 0.07717 & 0.09166 & 0.26186 & 0.44955 \\
Hand & 0.06384 & 0.09973 & 0.22999 & 0.29002 \\
\bottomrule
\end{tabular}

}
\vspace{-0.6cm}
\end{table}

\subsection{Age}
\label{subsec:age}
In this experiment, the data instances are grouped by the subject age (into buckets) with the genuine and impostor score distributions computed within the groups. Table \ref{table:agestats} shows the descriptive statistics of those score distributions.

\begin{table}[!ht]
\centering
\caption{Score distribution statistics by subject age}
\label{table:agestats}
\resizebox{\linewidth}{!}{
\begin{tabular}{lllllll}
\toprule
\multirow{2}{*}{Dataset} & \multirow{2}{*}{Algorithm} & \multirow{2}{*}{Age} & \multicolumn{2}{c}{Score $\mu$} & \multicolumn{2}{c}{Score $\sigma$} \\
& & & Genuine & Impostor & Genuine & Impostor \\\midrule
MMCBNU & LCNN & (0, 30) & 0.72681 & 0.26173 & 0.12008 & 0.08046 \\
& & (30, 45) & 0.71208 & 0.26273 & 0.12100 & 0.07735 \\
& & (45, 60) & 0.72994 & 0.25940 & 0.15319 & 0.07661 \\
& & (60, 80) & 0.74951 & 0.26902 & 0.09673 & 0.08338 \\
& LBP & (0, 30) & 0.84773 & 0.78782 & 0.01985 & 0.00805 \\
& & (30, 45) & 0.84817 & 0.78764 & 0.02005 & 0.00821 \\
& & (45, 60) & 0.85522 & 0.78854 & 0.02391 & 0.00740 \\
& & (60, 80) & 0.85705 & 0.78637 & 0.02181 & 0.00714 \\
& MC & (0, 30) & 0.27095 & 0.12056 & 0.04878 & 0.01734 \\
& & (30, 45) & 0.27476 & 0.12142 & 0.04635 & 0.01731 \\
& & (45, 60) & 0.28507 & 0.11776 & 0.06421 & 0.01648 \\
& & (60, 80) & 0.27749 & 0.11873 & 0.03664 & 0.01414 \\
& PC & (0, 30) & 0.40971 & 0.30607 & 0.02818 & 0.01651 \\
& & (30, 45) & 0.41318 & 0.30462 & 0.02556 & 0.01729 \\
& & (45, 60) & 0.41856 & 0.30536 & 0.03648 & 0.01512 \\
& & (60, 80) & 0.41563 & 0.30253 & 0.02026 & 0.01372 \\
& SIFT & (0, 30) & 0.38601 & 0.01520 & 0.16498 & 0.02444 \\
& & (30, 45) & 0.40078 & 0.01503 & 0.15826 & 0.02442 \\
& & (45, 60) & 0.46074 & 0.01360 & 0.19468 & 0.02238 \\
& & (60, 80) & 0.40609 & 0.00890 & 0.15790 & 0.01725 \\
\midrule
PLUS & LCNN & (0, 30) & 0.70530 & 0.24723 & 0.14055 & 0.08889 \\
& & (30, 45) & 0.70493 & 0.25356 & 0.13379 & 0.09098 \\
& & (45, 60) & 0.70513 & 0.27104 & 0.13314 & 0.09057 \\
& & (60, 80) & 0.69106 & 0.25425 & 0.14891 & 0.08597 \\
& LBP & (0, 30) & 0.82179 & 0.37210 & 0.15764 & 0.05956 \\
& & (30, 45) & 0.81593 & 0.37284 & 0.14697 & 0.05873 \\
& & (45, 60) & 0.75727 & 0.36913 & 0.17569 & 0.05507 \\
& & (60, 80) & 0.76103 & 0.37095 & 0.17469 & 0.06076 \\
& MC & (0, 30) & 0.28311 & 0.12303 & 0.03717 & 0.00735 \\
& & (30, 45) & 0.28255 & 0.12219 & 0.03647 & 0.00769 \\
& & (45, 60) & 0.28268 & 0.12447 & 0.04212 & 0.00770 \\
& & (60, 80) & 0.27906 & 0.12503 & 0.03901 & 0.00722 \\
& PC & (0, 30) & 0.43334 & 0.31615 & 0.02398 & 0.01348 \\
& & (30, 45) & 0.43270 & 0.31266 & 0.02275 & 0.01298 \\
& & (45, 60) & 0.42748 & 0.31316 & 0.02832 & 0.01336 \\
& & (60, 80) & 0.42937 & 0.31625 & 0.02424 & 0.01206 \\
& SIFT & (0, 30) & 0.46978 & 0.01702 & 0.14696 & 0.01886 \\
& & (30, 45) & 0.47233 & 0.01611 & 0.13525 & 0.01678 \\
& & (45, 60) & 0.44450 & 0.01227 & 0.15632 & 0.01360 \\
& & (60, 80) & 0.43647 & 0.01328 & 0.15149 & 0.01535 \\
\midrule
UTFVP & LCNN & (0, 30) & 0.69782 & 0.32213 & 0.10992 & 0.10830 \\
& & (30, 45) & 0.70565 & 0.33416 & 0.12312 & 0.11310 \\
& & (45, 60) & 0.68730 & 0.31387 & 0.13964 & 0.10498 \\
& LBP & (0, 30) & 0.84673 & 0.81045 & 0.01307 & 0.00475 \\
& & (30, 45) & 0.84676 & 0.80946 & 0.01522 & 0.00473 \\
& & (45, 60) & 0.84596 & 0.81029 & 0.01485 & 0.00435 \\
& MC & (0, 30) & 0.23783 & 0.11792 & 0.03854 & 0.00740 \\
& & (30, 45) & 0.24806 & 0.11696 & 0.03683 & 0.00667 \\
& & (45, 60) & 0.23632 & 0.11808 & 0.03878 & 0.00707 \\
& PC & (0, 30) & 0.40093 & 0.28823 & 0.02804 & 0.01072 \\
& & (30, 45) & 0.40530 & 0.29107 & 0.02974 & 0.00960 \\
& & (45, 60) & 0.40391 & 0.28678 & 0.02781 & 0.00953 \\
& SIFT & (0, 30) & 0.30731 & 0.00945 & 0.15074 & 0.01077 \\
& & (30, 45) & 0.35794 & 0.00984 & 0.17626 & 0.01092 \\
& & (45, 60) & 0.32461 & 0.00848 & 0.15849 & 0.00999 \\
\midrule
VERA & LBP & (0, 30) & 0.81264 & 0.78892 & 0.01086 & 0.00414 \\
& & (30, 45) & 0.81499 & 0.78846 & 0.00995 & 0.00413 \\
& & (45, 60) & 0.81130 & 0.78790 & 0.01243 & 0.00417 \\
& MC & (0, 30) & 0.23409 & 0.11333 & 0.04560 & 0.00945 \\
& & (30, 45) & 0.24782 & 0.11324 & 0.04133 & 0.00929 \\
& & (45, 60) & 0.25276 & 0.11431 & 0.04766 & 0.00964 \\
& PC & (0, 30) & 0.38283 & 0.29608 & 0.03144 & 0.01165 \\
& & (30, 45) & 0.39401 & 0.29441 & 0.02688 & 0.01053 \\
& & (45, 60) & 0.39404 & 0.29263 & 0.02745 & 0.01199 \\
& SIFT & (0, 30) & 0.21685 & 0.00603 & 0.14440 & 0.00856 \\
& & (30, 45) & 0.25496 & 0.00511 & 0.13332 & 0.00777 \\
& & (45, 60) & 0.25356 & 0.00506 & 0.15645 & 0.00768 \\
\bottomrule
\end{tabular}

}
\vspace{-0.6cm}
\end{table}

\subsection{Finger/Hand}
\label{subsec:fingerhand}
In this experiment, the data instances are grouped by the finger (index, middle, ring) or hand (left, right) with the genuine and impostor score distributions computed within the groups. Tables \ref{table:fingerstats} and \ref{table:handstats} show the descriptive statistics of those score distributions.

\begin{table}[!ht]
\centering
\caption{Score distribution statistics by subject finger}
\label{table:fingerstats}
\resizebox{\linewidth}{!}{
\begin{tabular}{lllllll}
\toprule
\multirow{2}{*}{Dataset} & \multirow{2}{*}{Algorithm} & \multirow{2}{*}{Finger} & \multicolumn{2}{c}{Score $\mu$} & \multicolumn{2}{c}{Score $\sigma$} \\
& & & Genuine & Impostor & Genuine & Impostor \\\midrule
MMCBNU & LCNN & Index & 0.73789 & 0.29660 & 0.11795 & 0.08182 \\
& & Middle & 0.73756 & 0.26873 & 0.11508 & 0.07969 \\
& & Ring & 0.69886 & 0.29819 & 0.12638 & 0.08600 \\
& LBP & Index & 0.85083 & 0.79090 & 0.01973 & 0.00675 \\
& & Middle & 0.85045 & 0.79195 & 0.01959 & 0.00632 \\
& & Ring & 0.84307 & 0.79134 & 0.02000 & 0.00708 \\
& MC & Index & 0.27596 & 0.12006 & 0.04771 & 0.01683 \\
& & Middle & 0.27706 & 0.12477 & 0.04816 & 0.01881 \\
& & Ring & 0.26312 & 0.12124 & 0.04960 & 0.01695 \\
& PC & Index & 0.41200 & 0.30462 & 0.02856 & 0.01614 \\
& & Middle & 0.41366 & 0.31002 & 0.02642 & 0.01815 \\
& & Ring & 0.40611 & 0.30702 & 0.02859 & 0.01608 \\
& SIFT & Index & 0.40806 & 0.01389 & 0.16277 & 0.02202 \\
& & Middle & 0.40803 & 0.01753 & 0.16304 & 0.02634 \\
& & Ring & 0.35635 & 0.01822 & 0.16490 & 0.02812 \\
\midrule
PLUS & LCNN & Index & 0.69603 & 0.31173 & 0.13513 & 0.09410 \\
& & Middle & 0.72436 & 0.32433 & 0.12724 & 0.09138 \\
& & Ring & 0.69037 & 0.31674 & 0.14844 & 0.09366 \\
& LBP & Index & 0.77026 & 0.39889 & 0.17601 & 0.06769 \\
& & Middle & 0.82733 & 0.42928 & 0.13097 & 0.06253 \\
& & Ring & 0.81040 & 0.40845 & 0.16954 & 0.06858 \\
& MC & Index & 0.27348 & 0.12309 & 0.04082 & 0.00786 \\
& & Middle & 0.28568 & 0.12384 & 0.03394 & 0.00713 \\
& & Ring & 0.28800 & 0.12297 & 0.03753 & 0.00763 \\
& PC & Index & 0.42761 & 0.31202 & 0.02737 & 0.01388 \\
& & Middle & 0.43428 & 0.31422 & 0.02249 & 0.01201 \\
& & Ring & 0.43326 & 0.32050 & 0.02292 & 0.01275 \\
& SIFT & Index & 0.43889 & 0.01861 & 0.15240 & 0.01762 \\
& & Middle & 0.47702 & 0.01313 & 0.13071 & 0.01350 \\
& & Ring & 0.47255 & 0.02040 & 0.15133 & 0.02180 \\
\midrule
UTFVP & LCNN & Index & 0.69280 & 0.35131 & 0.11088 & 0.11909 \\
& & Middle & 0.69936 & 0.36379 & 0.11254 & 0.10415 \\
& & Ring & 0.69923 & 0.36727 & 0.12042 & 0.11219 \\
& LBP & Index & 0.84579 & 0.81212 & 0.01321 & 0.00518 \\
& & Middle & 0.84737 & 0.81179 & 0.01281 & 0.00446 \\
& & Ring & 0.84677 & 0.81165 & 0.01421 & 0.00474 \\
& MC & Index & 0.23730 & 0.11875 & 0.03743 & 0.00743 \\
& & Middle & 0.23542 & 0.11804 & 0.03727 & 0.00737 \\
& & Ring & 0.24230 & 0.11776 & 0.04052 & 0.00730 \\
& PC & Index & 0.40019 & 0.28935 & 0.02829 & 0.01115 \\
& & Middle & 0.40176 & 0.28892 & 0.02687 & 0.01004 \\
& & Ring & 0.40274 & 0.28858 & 0.02922 & 0.01116 \\
& SIFT & Index & 0.30573 & 0.01048 & 0.15129 & 0.01137 \\
& & Middle & 0.31932 & 0.00907 & 0.14986 & 0.01025 \\
& & Ring & 0.31302 & 0.01048 & 0.16053 & 0.01153 \\
\midrule
VERA & LBP & Index & 0.81296 & 0.78861 & 0.01101 & 0.00417 \\
& MC & Index & 0.24056 & 0.11351 & 0.04569 & 0.00947 \\
& PC & Index & 0.38741 & 0.29499 & 0.03025 & 0.01156 \\
& SIFT & Index & 0.23220 & 0.00562 & 0.14523 & 0.00822 \\
\bottomrule
\end{tabular}

}
\vspace{-0.3cm}
\end{table}

\begin{table}[!ht]
\centering
\caption{Score distribution statistics by subject hand}
\label{table:handstats}
\resizebox{\linewidth}{!}{
\begin{tabular}{lllllll}
\toprule
\multirow{2}{*}{Dataset} & \multirow{2}{*}{Algorithm} & \multirow{2}{*}{Hand} & \multicolumn{2}{c}{Score $\mu$} & \multicolumn{2}{c}{Score $\sigma$} \\
& & & Genuine & Impostor & Genuine & Impostor \\\midrule
MMCBNU & LCNN & Left & 0.73241 & 0.27535 & 0.11214 & 0.08060 \\
& & Right & 0.72972 & 0.28659 & 0.10971 & 0.08007 \\
& LBP & Left & 0.84654 & 0.78811 & 0.02014 & 0.00803 \\
& & Right & 0.84969 & 0.78784 & 0.01993 & 0.00819 \\
& MC & Left & 0.26829 & 0.12023 & 0.04914 & 0.01738 \\
& & Right & 0.27581 & 0.12124 & 0.04838 & 0.01735 \\
& PC & Left & 0.40823 & 0.30551 & 0.02880 & 0.01697 \\
& & Right & 0.41295 & 0.30658 & 0.02710 & 0.01618 \\
& SIFT & Left & 0.37725 & 0.01505 & 0.16594 & 0.02398 \\
& & Right & 0.40438 & 0.01532 & 0.16370 & 0.02490 \\
\midrule
PLUS & LCNN & Left & 0.71348 & 0.25962 & 0.13772 & 0.09873 \\
& & Right & 0.69663 & 0.29911 & 0.13051 & 0.09379 \\
& LBP & Left & 0.79772 & 0.38143 & 0.16295 & 0.06064 \\
& & Right & 0.80837 & 0.37036 & 0.15992 & 0.06330 \\
& MC & Left & 0.28116 & 0.12307 & 0.03837 & 0.00744 \\
& & Right & 0.28375 & 0.12302 & 0.03763 & 0.00768 \\
& PC & Left & 0.43033 & 0.31581 & 0.02397 & 0.01204 \\
& & Right & 0.43320 & 0.31342 & 0.02495 & 0.01422 \\
& SIFT & Left & 0.45395 & 0.01591 & 0.14634 & 0.01645 \\
& & Right & 0.47223 & 0.01610 & 0.14504 & 0.01828 \\
\midrule
UTFVP & LCNN & Left & 0.69651 & 0.33332 & 0.11244 & 0.11739 \\
& & Right & 0.69369 & 0.35340 & 0.11329 & 0.10294 \\
& LBP & Left & 0.84549 & 0.81043 & 0.01359 & 0.00483 \\
& & Right & 0.84780 & 0.81076 & 0.01319 & 0.00476 \\
& MC & Left & 0.23586 & 0.11704 & 0.03855 & 0.00711 \\
& & Right & 0.24082 & 0.11883 & 0.03838 & 0.00752 \\
& PC & Left & 0.39946 & 0.28827 & 0.02852 & 0.01059 \\
& & Right & 0.40367 & 0.28838 & 0.02764 & 0.01052 \\
& SIFT & Left & 0.30254 & 0.00917 & 0.15592 & 0.01036 \\
& & Right & 0.32287 & 0.00966 & 0.15148 & 0.01107 \\
\midrule
VERA & LBP & Left & 0.81147 & 0.78964 & 0.01143 & 0.00402 \\
& & Right & 0.81446 & 0.78940 & 0.01036 & 0.00432 \\
& MC & Left & 0.23471 & 0.11385 & 0.04772 & 0.00959 \\
& & Right & 0.24641 & 0.11477 & 0.04277 & 0.00939 \\
& PC & Left & 0.38461 & 0.29604 & 0.03262 & 0.01128 \\
& & Right & 0.39022 & 0.29487 & 0.02740 & 0.01167 \\
& SIFT & Left & 0.21957 & 0.00583 & 0.14676 & 0.00826 \\
& & Right & 0.24482 & 0.00612 & 0.14257 & 0.00868 \\
\bottomrule
\end{tabular}

}
\end{table}

\subsection{Summary}
\label{subsec:summary}
In order to check whether the small differences reported in previous tables are statistically significant, the standard score ($Z$-score) is computed as shown in equation \ref{eq:zscore} (absolute value of the means is used, since only the magnitude of the difference is interesting in this case). Using $Z$-score is possible, since all-against-all comparisons were conducted and as such, the whole population of the comparison scores for each database/algorithm is known.

\begin{equation}
\label{eq:zscore}
Z = \frac{|\mu_{1} - \mu_{2}|}{\sqrt{\sigma^{2}_{1} - \sigma^{2}_2}}
\end{equation}

This computation is done for all the relevant pairs of score distributions for all the experiments. In other words, for each database and recognition algorithm, all permutations of the genuine and impostor distribution pairs are considered within the respective metadata attribute. Table \ref{table:zstats} shows the medians and maximums of the computed $Z$-scores. Overall, the $Z$-scores (medians) are very low and those of the outliers (maxima) are relatively low. In other words, statistically significant differences are not present for any of the score distribution pairs within their respective experiments (database, algorithm, and metadata attribute). The biometric performance evaluations (in verification and closed-set identification modes) have also not revealed any statistically significant differences.

\section{Conclusion}
\label{sec:conclusion}
As shown in the evaluation in section \ref{sec:evaluation}, statistically significant biases in score distributions \wrt the sex and age of the data subjects, as well as the chosen finger/hand \textbf{have not been detected} for the five fingervein recognition algorithms tested on the four datasets. Accordingly, no impact on the biometric performance in neither the biometric verification nor biometric identification mode has been discovered. The results thus indicate that various fundamentally different classes of fingervein recognition algorithms might be suitable for application irrespective of the tested meta-parameters. This also points to a potential advantage in certain application scenarios of the vascular characteristics in comparison to others (\eg face, see section \ref{sec:introduction}), for which potential biases have been reported in the literature. An obvious limitation of this work is the size of the used fingervein datasets -- unfortunately, no larger ones (with metadata present) are currently publicly available. Hence, an important avenue of future research in this area would be the acquisition of a more sizeable dataset and a validation of the scalability of those preliminary results. \vspace{-0.5cm}

\section*{Acknowledgements}
\label{sec:ackonwledgements}
\vspace{-0.1cm}
This research work has been funded by the German Federal Ministry of Education and Research
and the Hessen State Ministry for Higher Education, Research and the Arts within their joint support of the National Research Center for Applied Cybersecurity ATHENE, the LOEWE-3 BioBiDa Project (594/18-17), the FFG KIRAS project AUTFingerATM under grant No. 864785, and the FWF project "Advanced Methods and Applications for Fingervein Recognition" under grant No. P 32201-NBL.

\bibliographystyle{IEEEtran}
\bibliography{references}

\end{document}